\documentclass{article}

\usepackage{spconf}
\usepackage[T1]{fontenc}
\usepackage{lmodern}
\usepackage{amsmath,amsfonts,amssymb,mathtools,mathrsfs,amsthm}
\usepackage{thmtools,thm-restate}
\usepackage{cases,listings}
\usepackage{fancyvrb}
\usepackage[utf8]{inputenc} 
\usepackage[T1]{fontenc} 
\usepackage{hyperref}
\usepackage{url}
\usepackage{booktabs}
\usepackage{comment}
\usepackage{nicefrac}
\usepackage{microtype}
\usepackage{subcaption}
\usepackage{graphicx}
\usepackage{float}
\usepackage{multirow}
\usepackage{rotating}
\usepackage{tabu}
\usepackage{tikz-cd}
\usepackage[shortlabels]{enumitem}

\usepackage{algorithm}
\usepackage{algorithmic}

\newcommand{\ba}{\begin{array}}
\newcommand{\ea}{\end{array}}

\newcommand{\br}{\mathbb{R}}

\newcommand{\EE}{\mathbf E}

\newcommand{\FCal}{\mathcal{F}}

\newcommand{\CCal}{\mathcal{C}}

\newcommand{\PCal}{\mathcal{P}}

\newcommand{\XCal}{\mathcal{X}}

\newcommand{\ZCal}{\mathcal{Z}}

\newcommand{\indc}{\textbf{1}}
\newcommand{\ones}{\textbf{1}_d}

\newtheorem{theorem}{Theorem}[section]

\newtheorem{lemma}[theorem]{Lemma}
\newtheorem{remark}{Remark}
\newtheorem{proposition}{Proposition}

\title{Relaxed Wasserstein with Applications to GANs}
\name{Xin Guo, Johnny Hong, Tianyi Lin and Nan Yang}
\address{University of California, Berkeley \\ \{xinguo, jcyhong, darren{\_}lin, nanyang\}@berkeley.edu.}
\begin{document}
%\ninept
%
\maketitle
\begin{abstract}
Wasserstein Generative Adversarial Networks (WGANs) provide a versatile class of models, which have attracted great attention in various applications. However, this framework has two main drawbacks: (i) Wasserstein-1 (or Earth-Mover) distance is restrictive such that WGANs cannot always fit data geometry well; (ii) It is difficult to achieve fast training of WGANs. In this paper, we propose a new class of \textit{Relaxed Wasserstein} (RW) distances by generalizing Wasserstein-1 distance with Bregman cost functions. We show that RW distances achieve nice statistical properties while not sacrificing the computational tractability. Combined with the GANs framework, we develop Relaxed WGANs (RWGANs) which are not only statistically flexible but can be approximated efficiently using heuristic approaches. Experiments on real images demonstrate that the RWGAN with Kullback-Leibler (KL) cost function outperforms other competing approaches, e.g., WGANs, even with gradient penalty.  
\end{abstract}

\section{Introduction}\label{sec:intro}
Generative Adversarial Networks (GANs)~\cite{Goodfellow-2014-GAN} are a class of approaches for learning generative models based on game theory. They find various applications in image processing~\cite{Denton-2015-Deep, Reed-2016-Text, Yeh-2017-Semantic}, wireless communications~\cite{Shea-2016-Convolutional, Davaslioglu-2018-Generative} and signal processing~\cite{Li-2018-Speech, Eskimez-2019-Speech}. The GANs framework can be also easily modified with other loss functions and learning dynamics, triggering numerous variants~\cite{Mirza-2014-Conditional, Odena-2016-Conditional, Chen-2016-Infogan, Arjovsky-2017-Wasserstein, Mao-2017-Least, Zhao-2017-Energy, Mescheder-2017-Adversarial}. For an overview of the GANs theory and a bunch of relevant applications, we refer to the survey papers~\cite{Goodfellow-2016-GANs, Creswell-2018-Generative}. 

Despite the popularity of GANs, the training requires finding a Nash equilibrium of a nonconvex continuous game with high-dimensional parameters where the gradient-based algorithms fail to converge~\cite{Heusel-2017-Gans, Daskalakis-2018-Training}. There have been many attempts to mitigate this curse when using GANs, whether through minibatch discrimination and batch normalization~\cite{Saliman-2016-Improved, Brock-2018-Large}; averaging and extragradient~\cite{Mertikopoulos-2018-Optimistic} or by using the Wasserstein-1 distance~\cite{Arjovsky-2017-Wasserstein}. In particular, the dual representation of Wasserstein-1 distance~\cite{Villani-2008-Optimal} provides a theoretical foundation for reducing mode collapse and stabilizing learning process. The gradient penalty technique is also proposed for training Wasserstein GANs (WGANs)~\cite{Gulrajani-2017-Improved}. However, Wasserstein-1 distance is too restrictive to fit data geometry and WGAN training converges slowly even though it is more stable than other approaches.  

In this paper, we propose a new class of \textit{Relaxed Wasserstein} (RW) distances by generalizing Wasserstein-1 distance with Bregman cost functions\footnote{The Bregman functions are acknowledged for their superior performance in signal processing; see~\cite{Virtanen-2007-Monaural, Smaragdis-2014-Static}.}. Our new distances provide a unified mathematical framework for learning generative models with a good balance between model adaptability and computational tractability. Our contributions can be summarized as follows.  

First, we study the statistical behavior of RW distances. In particular, we prove that RW distances are dominated by the total variation (TV) distance and provide the non-asymptotic moment estimates for RW distances under certain conditions. Second, we show that RW distances permit a nice duality representation, which allows for gradient evaluation of RW distances. Based on this result and the asymmetric clipping heuristic method, we develop a gradient-based algorithm for RWGANs. Finally, we conduct experiments on real images. In the experiment, we select Kullback-Leibler (KL) cost function and the commonly-used DCGAN and MLP architectures. Experimental results demonstrate that our approach not only generates excellent samples but strikes a good balance between WGAN and WGAN with gradient penalty, denoted as WGANs(g). More specifically, our approach converges faster than WGAN and is more robust than WGAN(g) which is likely to fail in practice. Our approach is thus an effective alternative to WGAN and WGAN(g) for learning generative models on large-scale datasets. 

\textbf{Organization.} The rest of this paper is organized as follows. We introduce the basic setup and define the Relaxed Wasserstein (RW) distances in Section~\ref{sec:prelim}. We present the theoretical results as well as the gradient-based algorithm in Section~\ref{sec:result}. We report the numerical results on real images from four large-scale datasets in Section~\ref{sec:experiments} and conclude in Section~\ref{sec:conclusion}.

\textbf{Notations.} Throughout the paper, $\|\cdot\|$ refers to the Euclidean norm (in the corresponding vector space). We denote lowercase and uppercase letters as vectors and matrices. $x^\top$ and $\log(x)$ stand for the transpose and componentwise logarithm of $x$. We denote $\ones$ as a $d$-dimensional vector whose entries are all 1. A matrix $X$ is positive semidefinite if $X \succeq 0$ or positive definite if $X \succ 0$. For a set $\XCal$, we define its diameter by $D_\XCal = \max_{x_1,x_2\in\XCal} \|x_1-x_2\|$ and an indicator function over $\XCal$ by $\indc_{\XCal}$. For a function $f: \XCal \rightarrow \br$, we say it is $K$-Lipschitz for some constant $K>0$ if $|f(x)-f(y)| \leq K\|x-y\|$ holds for all $x, y \in \XCal$. 

\section{Relaxed Wasserstein distances}\label{sec:prelim}
Let $\XCal$ be compact and $\PCal(\XCal)$ be the set of all the probability distributions $\mu$ defined on $\XCal$ such that $\int_\XCal \|x\| d\mu(x) < +\infty$. For $\mu, \nu \in \PCal(\XCal)$, their Wasserstein-1 distance is defined by
\begin{equation}\label{def:Wasserstein}
W_1(\mu, \nu) = \inf_{\pi \in \Pi(\mu, \nu)} \int_{\XCal\times\XCal} \|x-y\| \; d\pi(x, y),
\end{equation}
where $\Pi(\mu, \nu)$ denotes the set of couplings (or joint distributions) between $\mu$ and $\nu$. The Wasserstein-1 distance is used in WGANs~\cite{Arjovsky-2017-Wasserstein} and has a dual representation~\cite{Villani-2008-Optimal} as follows, 
\begin{equation}\label{def:Wasserstein-dual}
W_1(\mu, \nu) = \sup_{f \in \FCal(\XCal)} \left\{\int_\XCal f \; d\mu(x) - \int_\XCal f \; d\nu(x)\right\}, 
\end{equation}
where $\FCal(\XCal)$ is the set of all the 1-Lipschitz functions defined on the set $\XCal$.  
\begin{remark}
Wasserstein-1 distance is indeed a metric: (i) $W_1(\mu, \nu) \geq 0$ and the equality holds if and only if $\mu=\nu$ almost everywhere, (ii) $W_1(\mu, \nu) = W_1(\nu, \mu)$, and (iii) $W_1(\mu, \nu) \leq W_1(\mu, \gamma) + W_1(\gamma, \nu)$ where $\gamma \in \PCal(\XCal)$ is also a probability distribution defined on $\XCal$.
\end{remark}
Another key ingredient in our paper is a class of cost functions based on the celebrated Bregman distances~\cite{Bregman-1967-Relaxation}. Specifically, let $x, y \in \XCal$, the Bregman cost function between them is defined by 
\begin{equation}\label{def:Bregman}
B_\phi(x, y) = \phi(x)-\phi(y)-\langle\nabla\phi(y), x-y\rangle, 
\end{equation}
where $\phi: \XCal \rightarrow \br$ is strictly convex and continuously differentiable. Examples of the function $\phi$ and the resulting Bregman cost functions are listed in Table~\ref{tab:example}. 
\begin{remark}
Bregman cost functions are well-defined for all $\phi$ since $B_\phi(x,y) \geq 0$ for all $x,y \in \XCal$ and the equality holds if and only if $x = y$. However, they are not metrics in general because of asymmetry.
\end{remark}
\begin{remark}
Bregman distances are well motivated from a statistical viewpoint. They are asymptotically equivalent to $f$-divergences under certain conditions~\cite{Pardo-2003-Asymptotic} and provide theoretical guarantee for $K$-means clustering~\cite{Banerjee-2005-Clustering}. Bregman distances are also unique loss functions where the conditional expectation is the optimal predictor~\cite{Banerjee-2005-Optimality}.
\end{remark}
\begin{table}[!t]\small
\caption{\footnotesize{Examples of the function $\phi$ and the resulting Bregman cost functions. Note that $A \succeq 0$ is positive semidefinite.}}\label{tab:example}\vspace*{-.5em}
\hspace*{-.5em}\begin{tabular}{|l|c|c|} \hline
& $\phi(x)$ & $B_\phi(x,y)$ \\ \hline
Euclidean & $\|x\|^2$ & $\|x-y\|^2$ \\ \hline
Mahalanobis & $x^\top Ax$ & $(x-y)^\top A(x-y)$ \\ \hline
Itakura-Saito & $-\sum_i \log(x_i)$ & $\sum_i (\frac{x_i}{y_i} -\log(\frac{x_i}{y_i})-1)$ \\ \hline
Kullback-Leibler & $\sum_i x_i\log(x_i)$ & $\sum_i (x_i\log(\frac{x_i}{y_i})-x_i+y_i)$ \\ \hline
\end{tabular}\vspace*{-1em}
\end{table}
We propose our new \textit{Relaxed Wasserstein} (RW) distances, which generalizes Wasserstein-1 distance with Bregman cost functions. Specifically, let $\mu, \nu \in \PCal(\XCal)$ and $\phi: \XCal \rightarrow \br$ be strictly convex and continuously differentiable, the relaxed Wasserstein distance (parameterized by $\phi$) between $\mu$ and $\nu$ is defined by 
\begin{equation}\label{def:relaxed-Wasserstein}
RW_\phi(\mu, \nu) = \inf_{\pi \in \Pi(\mu, \nu)} \int_{\XCal \times \XCal} B_\phi(x,y) \; d\pi(x, y). 
\end{equation}  
\begin{remark} 
RW distances are nonnegative for all $\phi$ since $RW_\phi(\mu, \nu) \geq 0$ for all $\mu, \nu \in \PCal(\XCal)$ and the equality holds if and only if $\mu=\nu$ almost everywhere. However, they are not metrics in general because of asymmetry. Since $\XCal$ is compact, $RW_\phi(\mu, \nu) < +\infty$ for all $\mu, \nu \in \PCal(\XCal)$. 
\end{remark}
\begin{remark}
RW distances include one important case: Wasserstein KL distance, which is the RW distance with $\phi(x) = \sum_i x_i\log(x_i)$. This distance is useful when $\XCal$ shares similar properties with the probability simplex.   
\end{remark}
\begin{remark}
RW distances extend Wasserstein-1 distance to a more general class by relaxing its symmetry. We demonstrate that such generalization resolves the restrictive essence of Wasserstein-1 distance which may not fit data geometry well, as encouraged by the success of Bregman distance in practice and our numerical results.   
\end{remark}

\section{Main Results}\label{sec:result}
Throughout this section, we let $\phi: \XCal \rightarrow \br$ be strictly convex and continuously differentiable and $\XCal$ be compact.

First, we show that RW distances are dominated by the total variation (TV) distance under certain conditions. Let $\mu, \nu \in \PCal(\XCal)$, their TV distance is defined by 
\begin{equation}\label{def:TV}
TV(\mu, \nu) = \sup_{A \subseteq \XCal} |\mu(A)-\nu(A)|.   
\end{equation}
We summarize the result in the following proposition. 
\begin{proposition}\label{Prop:dominated-TV} 
Let $\mu, \nu \in \PCal(\XCal)$, there exists a constant $L>0$ depending on $\phi$ and $\XCal$ such that
\begin{equation}
RW_\phi(\mu, \nu) \leq LD_\XCal TV(\mu, \nu). 
\end{equation}
\end{proposition}
\begin{proof}
Since $\phi$ is strictly convex and continuously differentiable and $\XCal$ is compact, $\phi$ is Lipschitz over the set $\XCal$. Therefore, we have $B_\phi(x,y) \leq L\|x-y\|$ for some constant $L \geq 0$. Using the definition of RW distances, we have $RW_\phi(\mu, \nu) \leq LW_1(\mu, \nu)$. Since $\XCal$ is compact,~\cite[Theorem~6.15]{Villani-2008-Optimal} implies that $W_1(\mu, \nu) \leq D_\XCal TV(\mu, \nu)$. Putting these pieces together yields the desired result. 
\end{proof}
We prove that RW distances achieve the nonasymptotic moment estimates under certain conditions in the following theorem. Our bounds are derived based on classical moment estimates for Wasserstein-1 distance~\cite[Theorem~3.1]{Lei-2020-Convergence}; see also~\cite{Fournier-2015-Rate, Weed-2019-Sharp, Singh-2018-Minimax, Lin-2021-Projection} for other results. 
\begin{proposition}\label{Prop:moment-estimate}
Let $\mu_\star \in \PCal(\XCal)$ with $\int_\XCal \|x\|^2 d\mu_\star(x) < +\infty$ and $\widehat{\mu}_n$ is an empirical distribution based on $n$ samples from $\mu_\star$. We also assume that the problem dimension $d>2$. Then we have  
\begin{equation}
\EE[RW_\phi(\widehat{\mu}_n, \mu_\star)] \leq cn^{-1/d} \textnormal{ for all } n \geq 1,   
\end{equation}
where $c$ is a constant independent of $n$ and $d$. 
\end{proposition}
\begin{proof}
Since $d > 2$ and $\int_\XCal \|x\|^2d\mu_\star(x) < +\infty$, we derive from~\cite[Theorem~3.1]{Lei-2020-Convergence} with $p=1$ and $q=2$ that
\begin{equation*}
\EE[W_1(\widehat{\mu}_n, \mu_\star)] \leq c'n^{-1/d} \textnormal{ for all } n \geq 1. 
\end{equation*}
Using the same reasoning from the proof of Proposition~\ref{Prop:dominated-TV}, we have $RW_\phi(\widehat{\mu}_n, \mu_\star) \leq LW_1(\widehat{\mu}_n, \mu_\star)$ for some constant $L>0$ which is independent of $n$ and $d$. Putting these pieces together yields the desired result. 
\end{proof}
%--------------------------------------------------------------------------
\begin{algorithm}[!t]
\caption{RWGANs.} \label{Alg:RWGANs}
\begin{algorithmic}                   
\REQUIRE $\alpha$: learning rate (0.0005); $c$: clipping range (0.005); $s$: scaling level (0.01); $m$: batch size (64); $n_d$: discriminator iteration number (5); $N_{\max}$: the maximum number of epoch.  
\REQUIRE $\theta_0, w_0$: generator and discriminator parameters
\FOR{$N=1,2,\ldots,N_{\max}$}
\FOR{$t=1,\ldots,n_d$}
\STATE Sample a batch of real data $\{x_i\}_{i=1}^m$ from $\mu_r$.
\STATE Sample a batch of prior samples $\{z_i\}_{i=1}^m$ from $\mu_Z$.
\STATE $g_w \leftarrow (1/m)\sum_{i=1}^m (\nabla_w f_w(x_i) - \nabla_w f_w(g_\theta(z_i)))$. 
\STATE $w\leftarrow w+\alpha\cdot\text{RMSProp}(w, g_w)$. 
\STATE $w\leftarrow \text{clip}(w, -s(\nabla\phi)^{-1}(-c), s(\nabla\phi)^{-1}(c))$. 
\ENDFOR
\STATE Sample a batch of prior samples $\{z_i\}_{i=1}^m$ from $\mu_Z$.
\STATE $g_\theta \leftarrow -(1/m)\sum_{i=1}^m \nabla_\theta f_w(g_\theta(z_i))$. 
\STATE $\theta \leftarrow \theta-\alpha\cdot\text{RMSProp}(\theta, g_\theta)$.
\ENDFOR
\end{algorithmic}
\end{algorithm}
%------------------------------------------------------------------
We present the dual representation of RW distances in the following theorem. The proof technique is similar to that of Kantorovich-Rubinstein duality theorem~\cite{Kantorovich-1958-Space, Edwards-2011-Kantorovich} 
\begin{theorem}\label{Theorem:duality} 
Let $\mu, \nu \in \PCal(\XCal)$ with $\int_\XCal \|x\| d\mu(x) < +\infty$ and $\int_\XCal \|x\| d\nu(x) < +\infty$. Then we have
\begin{equation}
RW_\phi(\mu, \nu) = \sup_{f \in \hat{\FCal}(\XCal)} \left\{\int_\XCal f \; d\mu(x) - \int_\XCal f \; d\nu(x)\right\},
\end{equation}
where $\FCal_l(\XCal) \subseteq \hat{\FCal}(\XCal) \subseteq \FCal_u(\XCal)$ with $\FCal_l(\XCal)=\{f: |f(x) - f(x')| \leq B_\phi(x,x')\}$ and $\FCal_u(\XCal) = \{f: |f(x) - f(x')| \leq \sup_{y \in \XCal} |B_\phi(x,y) - B_\phi(x',y)|\}$. 
\end{theorem}
\begin{proof} 
By the continuity of the functional $\int_\XCal f \; d\mu(x) - \int_\XCal f \; d\nu(x)$ with respect to $f$, it suffices to prove that 
\begin{align}
RW_\phi(\mu, \nu) \geq \sup_{f \in \FCal_l(\XCal)} \left\{\int_\XCal fd\mu(x) - \int_\XCal fd\nu(x)\right\}, \label{Eq:lower_RW_bound} \\
RW_\phi(\mu, \nu) \leq \sup_{f \in \FCal_u(\XCal)} \left\{\int_\XCal fd\mu(x) - \int_\XCal fd\nu(x)\right\}. \label{Eq:upper_RW_bound}
\end{align}
Using the same reasoning from the proof of Proposition~\ref{Prop:dominated-TV}, we have $RW_\phi(\mu, \nu) \leq LW_1(\mu, \nu)$ for some constant $L>0$. Since $\mu, \nu \in \PCal(\XCal)$ with $\int_\XCal \|x\| d\mu(x) < +\infty$ and $\int_\XCal \|x\| d\nu(x) < +\infty$, we have $W_1(\mu, \nu) < +\infty$. Putting these pieces together yields that $RW_\phi(\mu, \nu) < +\infty$.

For two functions $g_1:\XCal \rightarrow \br$ and $g_2:\XCal \rightarrow \br$, we define the function $g_1 \oplus g_2: \XCal \times \XCal \rightarrow \br$ by $(g_1 \oplus g_2)(x_1, x_2) = g_1(x_1)+g_2(x_2)$ for all $(x_1, x_2) \in \XCal \times \XCal$. We also denote $\CCal(\XCal)$ as the set of all the bounded measure functions. By~\cite[Theorem~5.10]{Villani-2008-Optimal}, we have
\begin{equation*}
RW_\phi(\mu, \nu) = \sup_{(g_1,g_2) \in \tilde{\FCal}(\XCal)} \left\{\int_\XCal g_1 d\mu(x) + \int_\XCal g_2 d\nu(x)\right\},  
\end{equation*}
where $\tilde{\FCal}(\XCal) = \{(g_1, g_2): g_1 \oplus g_2 \leq B_\phi, g_1, g_2 \in \CCal(\XCal)\}$. Now we are ready to prove Eq.~\eqref{Eq:lower_RW_bound} and~\eqref{Eq:upper_RW_bound}. 

\noindent \textbf{Proving Eq.~\eqref{Eq:lower_RW_bound}:} By the definition, $f \in \FCal_l(\XCal)$ implies that $(f, -f) \in \tilde{\FCal}(\XCal)$. Thus, we have
\begin{eqnarray*}
RW_\phi(\mu, \nu) & \geq & \sup_{(f,-f) \in \tilde{\FCal}(\XCal)} \left\{\int_\XCal fd\mu(x) - \int_\XCal fd\nu(x)\right\} \\
& \geq & \sup_{f \in \FCal_l(\XCal)} \left\{\int_\XCal fd\mu(x) - \int_\XCal fd\nu(x)\right\}.
\end{eqnarray*}
This implies that Eq.~\eqref{Eq:lower_RW_bound} holds true. 

\noindent \textbf{Proving Eq.~\eqref{Eq:upper_RW_bound}:} Let $\varepsilon>0$, there exists $g_1, g_2 \in \CCal(\XCal)$ with $g_1 \oplus g_2 \leq B_\phi$ such that  
\begin{equation*}
RW_\phi(\mu, \nu) - \varepsilon \leq \int_\XCal g_1 \; d\mu(x) + \int_\XCal g_2 \; d\nu(x).
\end{equation*}
Now we construct a function $f \in \FCal_u(\XCal)$ out of $g_1$ and $g_2$. More specifically, we first define the function $f: \XCal \rightarrow \br$ by $f(x) = \inf_{y \in \XCal} \{B_\phi(x, y)-g_2(y)\}$ for all $x \in \XCal$. Since $g_2$ is bounded and $\XCal$ is compact, the function $f$ is well-defined. Moreover, for any $x, x' \in \XCal$, we have  
\begin{eqnarray*}
& & |f(x) - f(x')| \\
& = & \left|\inf_{y \in \XCal} \{B_\phi(x, y)-g_2(y)\} - \inf_{y \in \XCal} \{B_\phi(x', y)-g_2(y)\}\right| \\
& \leq & \sup_{y \in \XCal} |B_\phi(x,y) - B_\phi(x',y)|.  
\end{eqnarray*}
Furthermore, since $g_1 \oplus g_2 \leq B_\phi$, we have 
\begin{equation*}
g_1(x) \leq \inf_{y \in \XCal} \{B_\phi(x, y)-g_2(y)\} \leq B_\phi(x, x) - g_2(x) = -g_2(x). 
\end{equation*}
By the definition of $f$, we have $g_1 \leq f$ and $g_2 \leq -f$. Therefore, we conclude that 
\begin{eqnarray*}
RW_\phi(\mu, \nu) - \varepsilon & \leq & \int_\XCal f \; d\mu(x) - \int_\XCal f \; d\nu(x) \\ 
& \leq & \sup_{f \in \FCal_u(\XCal)} \left\{\int_\XCal f \; d\mu(x) - \int_\XCal f \; d\nu(x)\right\}. 
\end{eqnarray*}
Letting $\varepsilon \rightarrow 0$ yields that Eq.~\eqref{Eq:upper_RW_bound} holds true. 
\end{proof}
\begin{remark}
Our theorem is valid since $\FCal_l(\XCal) \subseteq \FCal_u(\XCal)$ holds true in general. However, the converse only holds true under additional conditions. For example, if the Bregman cost function is a metric, we can prove $\FCal_l(\XCal) = \FCal_u(\XCal)$ using the triangle inequality and our theorem exactly recovers the Kantorovich-Rubinstein duality theorem. 
\end{remark}
The goal of generative modeling is to estimate the unknown probability distribution $\mu_r$ using parametric probability distributions $\mu_\theta$. Indeed, we draw $z \in \ZCal$ from $\mu_Z$ and define $g_\theta: \ZCal \to \XCal$. Then $\mu_\theta$ is defined as the probability distribution of $g_\theta(z)$. 

In what follows, we consider this setting and focus on the computational aspect of $RW_\phi(\mu_r, \mu_\theta)$. 
\begin{theorem}\label{Theorem:computation-RW}
$RW_\phi(\mu_r, \mu_\theta)$ is \textbf{continuous} in $\theta$ if $g_\theta$ is continuous in $\theta$. $RW_\phi(\mu_r, \mu_\theta)$ is \textbf{differentiable} almost everywhere if $\nabla\phi$ is locally Lipschitz and $g_\theta$ is locally Lipschitz with a constant $L(\theta, z)$ satisfying $\EE_Z[L(\theta,Z)]<\infty$. If Theorem~\ref{Theorem:duality} also holds true, there exists a function $f \in \hat{\FCal}(\XCal)$ such that $\nabla_\theta [RW_\phi(\mu_r, \mu_\theta)] = -\EE_Z[\nabla_\theta f(g_\theta(Z)))]$ where $\hat{\FCal}(\XCal)$ is defined in Theorem~\ref{Theorem:duality}. 
\end{theorem}
Before the proof of Theorem~\ref{Theorem:computation-RW}, we present a technical lemma which is important to subsequent analysis. 
\begin{lemma}\label{Lemma:triangle-RW}
Let $\theta, \theta'$ be two parameters and $\mu_Z$ be the fixed probability distribution of the latent variable under the framework of generative modeling, we have
\begin{eqnarray*}
|RW_\phi(\mu_r, \mu_\theta) - RW_\phi(\mu_r, \mu_{\theta'})| & \leq & LW_1(\mu_\theta, \mu_{\theta'}) \\
& & \hspace{-12em} + D_\XCal\EE_Z[\|\nabla\phi(g_\theta(Z))-\nabla\phi(g_{\theta'}(Z))\|]. 
\end{eqnarray*}
where $L>0$ is a constant only depending on $\XCal$ and $\phi$. 
\end{lemma}
\begin{proof}
Let $(X_1, X_2)$ be an optimal coupling of $(\mu_r, \mu_\theta)$ and $(Z_2, Z_3)$ be an optimal coupling of $(\mu_\theta, \mu_{\theta'})$. By the gluing Lemma~\cite{Armstrong-2013-Basic}, there exist random variables $(X'_1, X'_2, X'_3)$ with $(X_1, X_2)=(X'_1, X'_2)$ and $(Z_2, Z_3)=(X'_2, X'_3)$ in distribution. Thus, $(X'_1, X'_3)$ is a coupling of $(\mu_r, \mu_{\theta'})$ and we have 
\begin{equation*}
RW_\phi(\mu_r, \mu_{\theta'}) \leq \EE[B_\phi(X'_1, X'_3)]. 
\end{equation*}
By definition, we have
\begin{eqnarray*}
B_\phi(X'_1, X'_3) & \leq & B_\phi(X'_1, X'_2) + B_\phi(X'_2, X'_3) \\ 
& & \hspace{-5em} + \|\nabla\phi(X'_2)-\nabla\phi(X'_3)\|\|X_1'-X_2'\|. 
\end{eqnarray*}
Putting these pieces together yields that 
\begin{eqnarray*}
RW_\phi(\mu_r, \mu_{\theta'}) & \leq & RW_\phi(\mu_r, \mu_\theta) + RW_\phi(\mu_\theta, \mu_{\theta'}) \\ 
& & \hspace{-5em} + \EE[\|\nabla\phi(X'_2)-\nabla\phi(X'_3)\|\|X_1'-X_2'\|]
\end{eqnarray*}
Since $\XCal$ is compact, we have 
\begin{eqnarray*}
RW_\phi(\mu_r, \mu_{\theta'}) & \leq & RW_\phi(\mu_r, \mu_\theta) + RW_\phi(\mu_\theta, \mu_{\theta'}) \\ 
& & \hspace{-5em} + D_\XCal\EE[\|\nabla\phi(X'_2)-\nabla\phi(X'_3)\|]. 
\end{eqnarray*}
Using the same reasoning from the proof of Proposition~\ref{Prop:dominated-TV}, we have $RW_\phi(\mu_\theta, \mu_{\theta'}) \leq LW_1(\mu_\theta, \mu_{\theta'})$ for some constant $L>0$. Putting these pieces together yields that 
\begin{eqnarray*}
& & RW_\phi(\mu_r, \mu_{\theta'}) - RW_\phi(\mu_r, \mu_\theta) \\ 
& \leq & LW_1(\mu_\theta, \mu_{\theta'}) + D_\XCal\EE[\|\nabla\phi(X'_2)-\nabla\phi(X'_3)\|]. 
\end{eqnarray*}
Since the right-hand side is symmetry with respect to $(\mu_\theta, \mu_{\theta'})$, we exchange $\mu_\theta$ and $\mu_{\theta'}$ to obtain that 
\begin{eqnarray*}
& & RW_\phi(\mu_r, \mu_\theta) - RW_\phi(\mu_r, \mu_{\theta'}) \\ 
& \leq & LW_1(\mu_\theta, \mu_{\theta'}) + D_\XCal\EE[\|\nabla\phi(X'_2)-\nabla\phi(X'_3)\|]. 
\end{eqnarray*}
By definition of $(X'_2, X'_3)$, we have $X_2' \sim \mu_\theta$ and $X'_3 \sim \mu_{\theta'}$. By the definition of $\mu_\theta$ and $\mu_{\theta'}$, we have
\begin{equation*}
\EE[\|\nabla\phi(X'_2)-\nabla\phi(X'_3)\|] = \EE_Z[\|\nabla\phi(g_\theta(Z))-\nabla\phi(g_{\theta'}(Z))\|]. 
\end{equation*}
Putting these pieces together yields the desired result. 
\end{proof}

\noindent\textbf{Proof of Theorem~\ref{Theorem:computation-RW}.} Our proof is based the techniques used for proving~\cite[Theorem~1 and~3]{Arjovsky-2017-Wasserstein}. 

First, by the definition of RW distances, we have $W_1(\mu_\theta, \mu_{\theta'}) \leq \EE_Z[\|g_\theta(Z)-g_{\theta'}(Z)\|]$. Since $g_\theta$ is continuous in $\theta$, then $g_\theta(z) \rightarrow g_{\theta'}(z)$ for all $z \in \ZCal$ as $\theta \rightarrow \theta'$. Thus, $\|g_\theta-g_{\theta'}\| \rightarrow 0$ pointwise as a function of $z$. Since $\XCal$ is compact, $\|g_\theta(z)-g_{\theta'}(z)\| \leq D_\XCal$ for all $\theta$ and $z \in \ZCal$ uniformly. By the bounded convergence theorem, we have
\begin{equation*}
W_1(\mu_\theta, \mu_{\theta'}) \leq \EE_Z[\|g_\theta(Z)-g_{\theta'}(Z)\|] \rightarrow 0 \textnormal{ as } \theta \rightarrow \theta'. 
\end{equation*}
Using the same argument and the continuity of $\nabla\phi$, we have $\EE_Z[\|\nabla\phi(g_\theta(Z))-\nabla\phi(g_{\theta'}(Z))\|] \rightarrow 0$ as $\theta \rightarrow \theta'$. Putting these pieces with Lemma~\ref{Lemma:triangle-RW} yields that 
\begin{equation*}
|RW_\phi(\mu_r, \mu_\theta) - RW_\phi(\mu_r, \mu_{\theta'})| \rightarrow 0 \textnormal{ as } \theta \rightarrow \theta'. 
\end{equation*}
This implies the continuity of $RW_\phi(\mu_r, \mu_\theta)$. 

Second, let $g_\theta$ be locally Lipschitz in $\theta$. Then, for a given pair $(\theta, z)$, there is a constant $L(\theta, z)$ and an open set $U$ such that $(\theta, z) \in U$, such that for every $(\theta', z') \in U$, we have
\begin{equation*}
\|g_\theta(z) - g_{\theta'}(z')\| \leq L(\theta, z)(\|\theta-\theta'\|+\|z-z'\|). 
\end{equation*}
By letting $(\theta',z') \in U$ with $z'=z$ and taking expectations with respect to $Z$, we have
\begin{equation*}
\EE_Z[\|g_\theta(Z) - g_{\theta'}(Z)\|] \leq \|\theta-\theta'\|\EE_Z[L(\theta, Z)]. 
\end{equation*}
Therefore, we can define $U_\theta=\{\theta': (\theta',z) \in U\}$. Since $U$ is open, $U_\theta$ is open. Also, $L(\theta)=\EE_Z[L(\theta, Z)]<+\infty$. Therefore, we have
\begin{equation*}
W_1(\mu_\theta, \mu_{\theta'}) \leq L(\theta)\|\theta-\theta'\|. 
\end{equation*}
Since $\nabla\phi$ is locally Lipschitz and $\XCal$ is compact, we have
\begin{equation*}
\EE_Z[\|\nabla\phi(g_\theta(Z))-\nabla\phi(g_{\theta'}(Z))\|] \leq L'\EE_Z[\|g_\theta(Z)-g_{\theta'}(Z)\|]. 
\end{equation*}
for some constants $L'>0$ only depending on $\XCal$ and $\phi$. Putting these pieces with Lemma~\ref{Lemma:triangle-RW} yields that
\begin{equation*}
|RW_\phi(\mu_r, \mu_\theta) - RW_\phi(\mu_r, \mu_{\theta'})| \leq (L+L'D_\XCal)L(\theta)\|\theta-\theta'\|. 
\end{equation*}
for all $\theta' \in U_\theta$, meaning that $RW_\phi(\mu_r, \mu_\theta)$ is locally Lipschitz in $\theta$. Using the celebrated Rademacher’s theorem (see~\cite[Theorem~3.1.6]{Federer-2014-Geometric} for example), we conclude that $RW_\phi(\mu_r, \mu_\theta)$ is differentiable almost everywhere. 

Finally, we define $V(\tilde{f}, \theta)$ by 
\begin{equation*}
V(\tilde{f}, \theta) = \int_\XCal \tilde{f} d\mu_r(x) - \int_\XCal \tilde{f} d\mu_\theta(x),  
\end{equation*}
where $\tilde{f}$ lies in the set $\hat{\FCal}(\XCal)$ and $\theta$ is a parameter. Since $\XCal$ is compact and Theorem~\ref{Theorem:duality} holds true, there is a function $f \in \hat{\FCal}(\XCal)$ such that $RW_\phi(\mu_r, \mu_\theta) = \sup_{\tilde{f} \in \hat{\FCal}(\XCal)} V(\tilde{f}, \theta) = V(f, \theta)$. Let us define $X^\circ(\theta)=\{f \in \hat{\FCal}(\XCal): V(f, \theta) = RW_\phi(\mu_r, \mu_\theta)\}$. Using the previous arguments, we obtain that $X^\circ(\theta)$ is nonempty. By a simple envelope theorem (see~\cite[Theorem~1]{Milgrom-2002-Envelope} for example) that
\begin{equation*}
\nabla_\theta [RW_\phi(\mu_r, \mu_\theta)] = \nabla_\theta V(f, \theta), 
\end{equation*}
for any $f \in X^\circ(\XCal)$ when both terms are well-defined. In such case, we further have
\begin{equation*}
\nabla_\theta V(f, \theta) = \nabla_\theta\left[\int_\XCal f d\mu_r(x) - \int_\XCal f d\mu_\theta(x)\right]. 
\end{equation*}
It is clear that the first term does not depend on $\theta$. This together with the relationship between $\mu_\theta$ and $Z$ under the framework of generative modeling yields that 
\begin{equation*}
\nabla_\theta V(f, \theta) = -\nabla_\theta\left[\int_\XCal f d\mu_\theta(x)\right] = -\nabla_\theta[\EE_Z[f(g_\theta(Z))]]. 
\end{equation*}
under the condition that the first and last terms are well-defined. Then it suffices to show that 
\begin{equation*}
\nabla_\theta[\EE_Z[f(g_\theta(Z))]] = \EE_Z[\nabla_\theta f(g_\theta(Z)))],  
\end{equation*}
when the right-hand side is defined. Since $f \in \hat{\FCal}(\XCal)$, we have $f \in \FCal_u(\XCal)$. By the definition of $\FCal_u(\XCal)$, we have 
\begin{equation*}
|f(x) - f(x')| \leq \sup_{y \in \XCal} |B_\phi(x,y) - B_\phi(x',y)|
\end{equation*}
for all $x, x' \in \XCal$. By the compactness of $\XCal$, we have $\sup_{y \in \XCal} |B_\phi(x,y) - B_\phi(x',y)| \leq L''\|x-x'\|$ for some constants $L''>0$, meaning that $f$ is also Lipschitz. 

The remaining proof is based on Radamacher’s theorem, Fubini's theorem and the dominated convergence theorem, and is exactly the same as that used for proving~\cite[Theorem~3]{Arjovsky-2017-Wasserstein}. This completes the proof.  
\begin{remark}
Together with~\cite[Corollary~1]{Arjovsky-2017-Wasserstein}, Theorem~\ref{Theorem:computation-RW} provides the strong theoretical guarantee for learning by minimizing the RW distances with neural networks. We refer the interested readers to~\cite{Arjovsky-2017-Wasserstein} for the details. 
\end{remark}
\textbf{RWGANs.} We define a random variable $Z$ with a fixed distribution $\mu_Z$ and draw a sample from $\mu_\theta$ by passing it through $g_\theta: \ZCal \to \XCal$. Then one approximates $\mu_r$ using $\mu_\theta$ by adapting $\theta$ to minimize RW distances between $\mu_r$ and $\mu_\theta$. In our experiment, we consider KL cost function where $\phi(x)=\sum_i x_i\log(x_i)$. Such choice is based on the observation that KL distance can fit different type of data geometry well in clustering tasks~\cite{Banerjee-2005-Clustering}.

We present a gradient-based algorithm\footnote{The value inside the bracket is default choice in our experiment.} for RWGANs; see Algorithm \ref{Alg:RWGANs}. While back propagation is used to train generator and discriminator and update the parameters once in the generator and $n_{critic}$ times in the discriminator, our framework is different from WGANs~\cite{Arjovsky-2017-Wasserstein}. Indeed, we use asymmetric clipping based on $\phi$ to guarantee that $f_w \in \hat{\FCal}(\XCal)$ while WGANs use symmetric clipping such that $f_w \in \FCal(\XCal)$. We also adopt RMSProp~\cite{Tieleman-2012-Lecture} with large stepsize which works well in practice.
\begin{table}[!t]
\caption{\footnotesize{Inception scores (IS) obtained by running RWGAN, WGAN and WGAN(g). For \textsc{cifar10}, ``begin" and ``end" refer to IS averaged over first 5 and last 10 epochs. For \textsc{imagenet}, ``begin" and ``end" refer to IS averaged over first 3 and last 5 epochs.}}\label{Tab:IS}\vspace*{-.5em}
\centering
\begin{tabular}{|c|c|c|c|c|c|} \hline
& \multirow{2}{*}{\footnotesize{Method}} & \multicolumn{2}{c|}{\textsc{cifar10}} & \multicolumn{2}{|c|}{\textsc{imagenet}} \\ \cline{3-6} 
& & begin & end & begin & end \\ \hline
\multirow{3}{*}{\footnotesize{DCGAN}} & \footnotesize{RWGAN} & \textbf{1.86} & 2.40 & 2.04 & 2.70 \\ 
& \footnotesize{WGAN} & 1.63 & \textbf{2.42} & 2.21 & \textbf{2.80} \\ 
& \footnotesize{WGAN(g)} & 1.73 & 2.37 & \textbf{2.27} & 2.73 \\ \hline
\multirow{3}{*}{\footnotesize{MLP}} & \footnotesize{RWGAN} & \textbf{1.31} & 2.17 & \textbf{2.00} & \textbf{2.48} \\ 
& \footnotesize{WGAN} & 1.28 & 1.90 & 1.74 & 2.23 \\ 
& \footnotesize{WGAN(g)} & 1.27 & \textbf{2.22} & 1.88 & 2.34 \\ \hline
\end{tabular}\vspace*{-1em}
\end{table}
\begin{figure*}[!t]\small
\hspace*{-5.2em}\begin{tabular}{cccccc||cccccc}
\multicolumn{6}{c}{\textsc{mnist}} & \multicolumn{6}{c}{\textsc{fashion-mnist}} \\ \hline
\footnotesize{RWGAN} & \footnotesize{WGAN} & \footnotesize{CGAN} & \footnotesize{InfoGAN} & \footnotesize{GAN} & \footnotesize{ACGAN} & \footnotesize{RWGAN} & \footnotesize{WGAN} & \footnotesize{CGAN} & \footnotesize{InfoGAN} & \footnotesize{GAN} & \footnotesize{ACGAN} \\
\begin{minipage}{0.075\textwidth}
\includegraphics[width=1\textwidth, height=1.3cm]{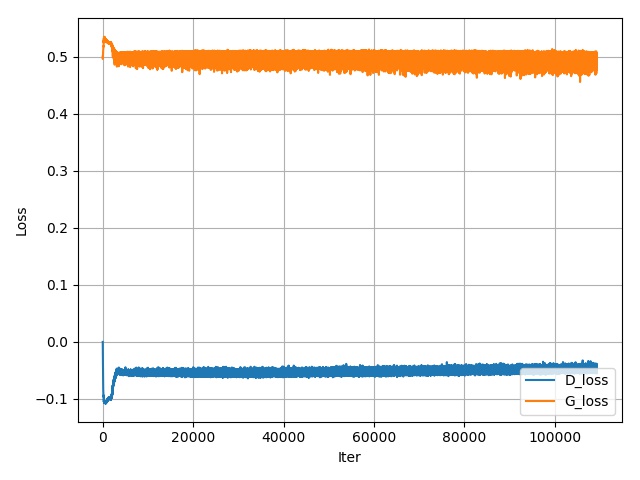}
\end{minipage}
& \begin{minipage}{0.075\textwidth}
\includegraphics[width=1\textwidth, height=1.3cm]{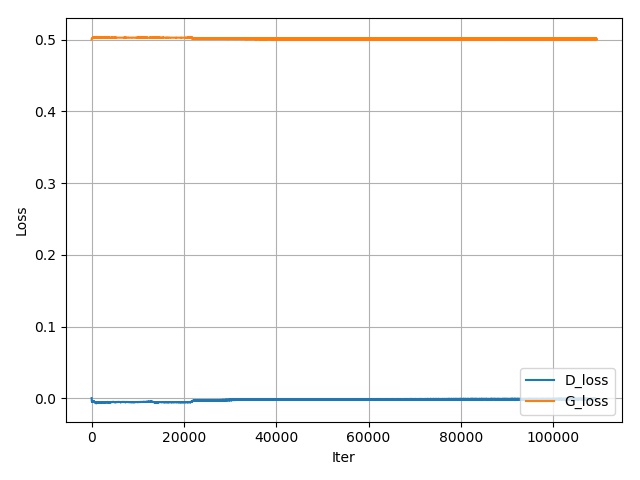}
\end{minipage}
& \begin{minipage}{0.075\textwidth}
\includegraphics[width=1\textwidth, height=1.3cm]{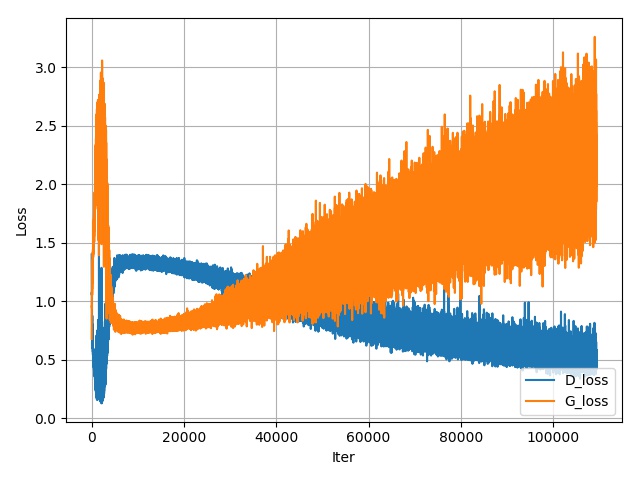}
\end{minipage}
& \begin{minipage}{0.075\textwidth}
\includegraphics[width=1\textwidth, height=1.3cm]{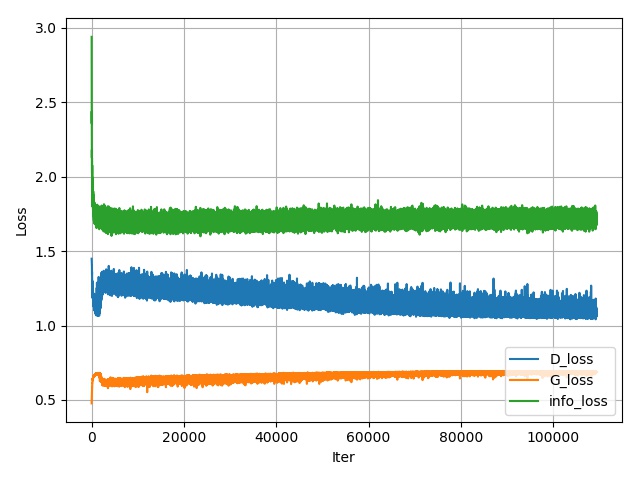}
\end{minipage}
& \begin{minipage}{0.075\textwidth}
\includegraphics[width=1\textwidth, height=1.3cm]{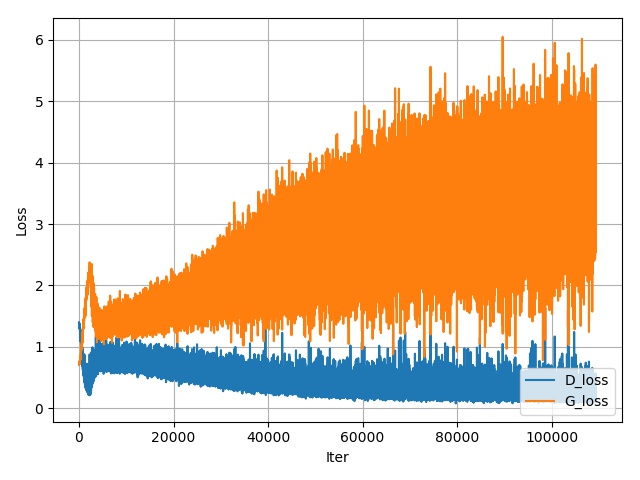}
\end{minipage}
& \begin{minipage}{0.075\textwidth}
\includegraphics[width=1\textwidth, height=1.3cm]{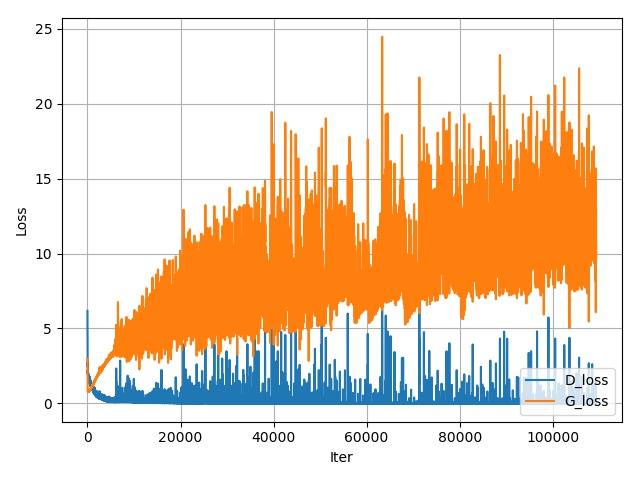}
\end{minipage}
& \begin{minipage}{0.075\textwidth}
\includegraphics[width=1\textwidth, height=1.3cm]{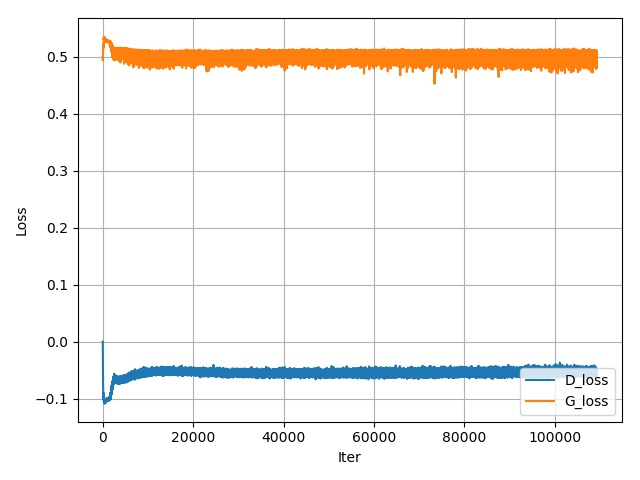}
\end{minipage}
& \begin{minipage}{0.075\textwidth}
\includegraphics[width=1\textwidth, height=1.3cm]{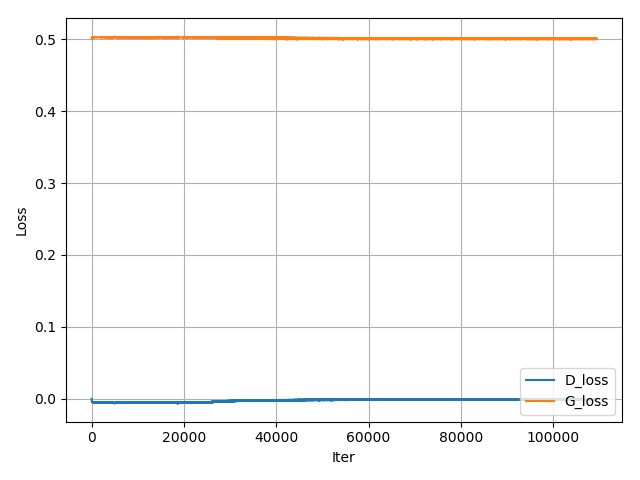}
\end{minipage}
& \begin{minipage}{0.075\textwidth}
\includegraphics[width=1\textwidth, height=1.3cm]{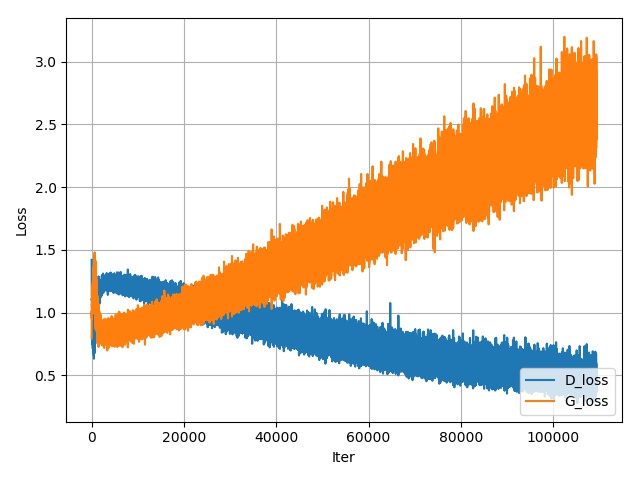}
\end{minipage}
& \begin{minipage}{0.075\textwidth}
\includegraphics[width=1\textwidth, height=1.3cm]{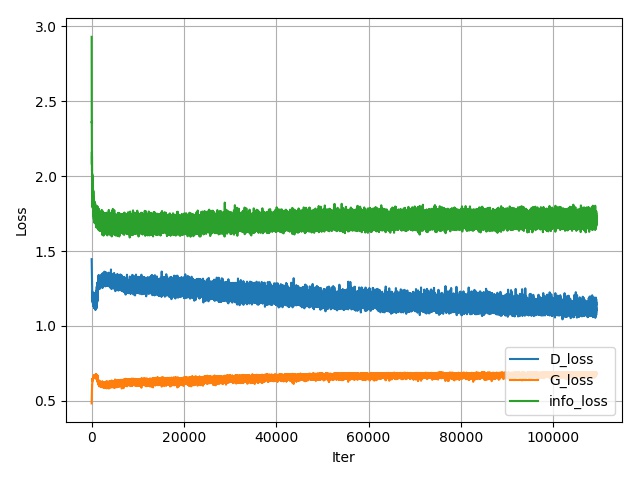}
\end{minipage}
& \begin{minipage}{0.075\textwidth}
\includegraphics[width=1\textwidth, height=1.3cm]{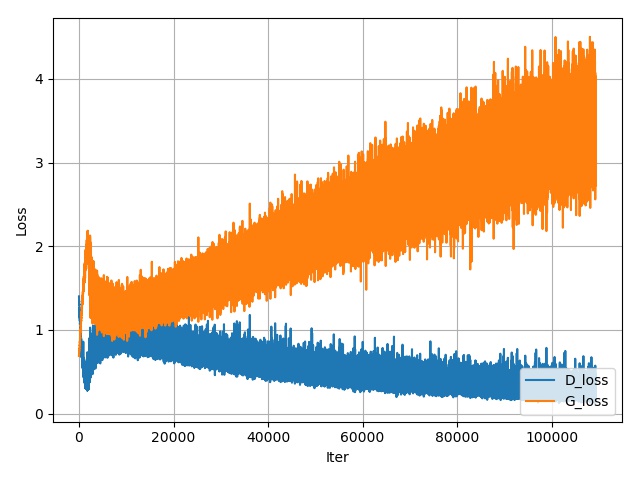}
\end{minipage}
& \begin{minipage}{0.075\textwidth}
\includegraphics[width=1\textwidth, height=1.3cm]{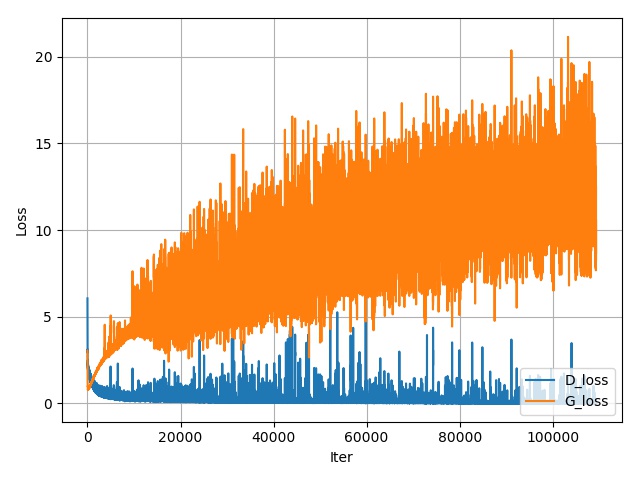}
\end{minipage}
\end{tabular}
\vspace*{-1em}\caption{\small{Training curves by running all approaches. Generator and discriminator losses are plotted in orange and blue lines.}}\label{Fig:train_curve_mnist}\vspace*{-1.2em}
\end{figure*}
\begin{figure*}[!t]\small
\hspace*{-5.2em}\begin{tabular}{cccccc||cccccc}
\multicolumn{6}{c}{\textsc{mnist}} & \multicolumn{6}{c}{\textsc{fashion-mnist}} \\ \hline
\footnotesize{RWGAN} & \footnotesize{WGAN} & \footnotesize{CGAN} & \footnotesize{InfoGAN} & \footnotesize{GAN} & \footnotesize{ACGAN} & \footnotesize{RWGAN} & \footnotesize{WGAN} & \footnotesize{CGAN} & \footnotesize{InfoGAN} & \footnotesize{GAN} & \footnotesize{ACGAN} \\
\begin{minipage}{0.075\textwidth}
\includegraphics[width=1\textwidth, height=1.3cm]{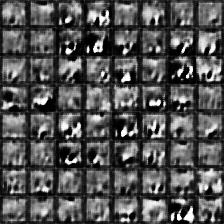}
\end{minipage}
& \begin{minipage}{0.075\textwidth}
\includegraphics[width=1\textwidth, height=1.3cm]{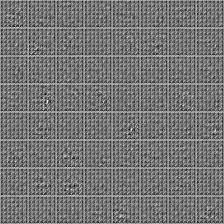}
\end{minipage}
& \begin{minipage}{0.075\textwidth}
\includegraphics[width=1\textwidth, height=1.3cm]{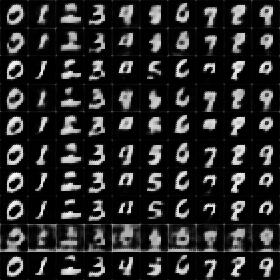}
\end{minipage}
& \begin{minipage}{0.075\textwidth}
\includegraphics[width=1\textwidth, height=1.3cm]{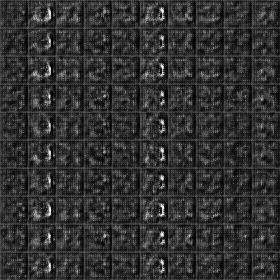}
\end{minipage}
& \begin{minipage}{0.075\textwidth}
\includegraphics[width=1\textwidth, height=1.3cm]{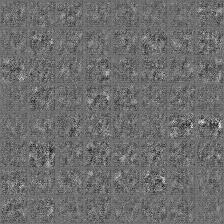}
\end{minipage}
& \begin{minipage}{0.075\textwidth}
\includegraphics[width=1\textwidth, height=1.3cm]{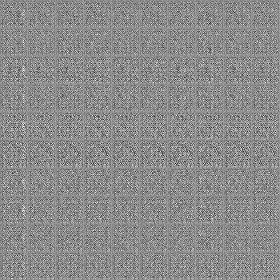}
\end{minipage}
& \begin{minipage}{0.075\textwidth}
\includegraphics[width=1\textwidth, height=1.3cm]{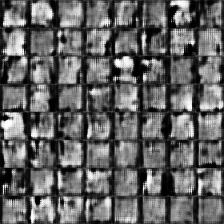}
\end{minipage}
& \begin{minipage}{0.075\textwidth}
\includegraphics[width=1\textwidth, height=1.3cm]{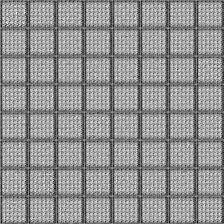}
\end{minipage}
& \begin{minipage}{0.075\textwidth}
\includegraphics[width=1\textwidth, height=1.3cm]{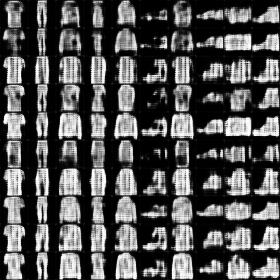}
\end{minipage}
& \begin{minipage}{0.075\textwidth}
\includegraphics[width=1\textwidth, height=1.3cm]{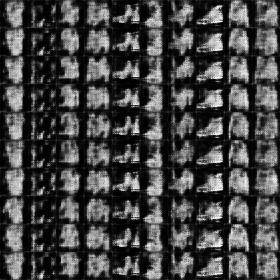}
\end{minipage}
& \begin{minipage}{0.075\textwidth}
\includegraphics[width=1\textwidth, height=1.3cm]{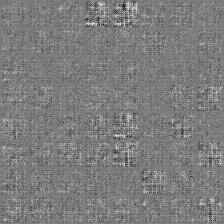}
\end{minipage}
& \begin{minipage}{0.075\textwidth}
\includegraphics[width=1\textwidth, height=1.3cm]{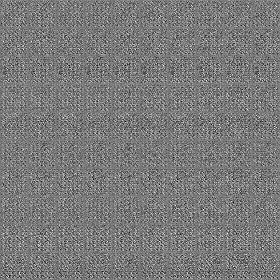}
\end{minipage} \\ \hline \hline
\begin{minipage}{0.075\textwidth}
\includegraphics[width=1\textwidth, height=1.3cm]{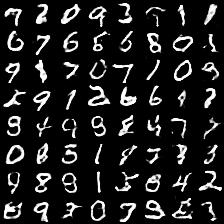}
\end{minipage}
& \begin{minipage}{0.075\textwidth}
\includegraphics[width=1\textwidth, height=1.3cm]{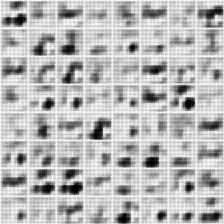}
\end{minipage}
& \begin{minipage}{0.075\textwidth}
\includegraphics[width=1\textwidth, height=1.3cm]{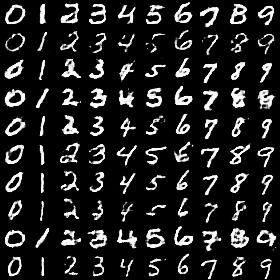}
\end{minipage}
& \begin{minipage}{0.075\textwidth}
\includegraphics[width=1\textwidth, height=1.3cm]{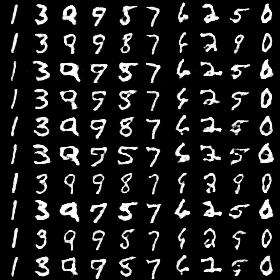}
\end{minipage}
& \begin{minipage}{0.075\textwidth}
\includegraphics[width=1\textwidth, height=1.3cm]{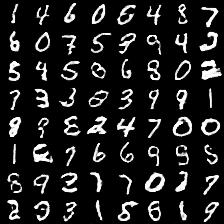}
\end{minipage}
& \begin{minipage}{0.075\textwidth}
\includegraphics[width=1\textwidth, height=1.3cm]{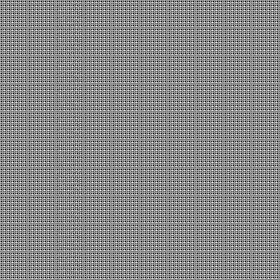}
\end{minipage}
& \begin{minipage}{0.075\textwidth}
\includegraphics[width=1\textwidth, height=1.3cm]{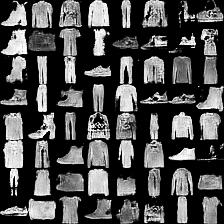}
\end{minipage}
& \begin{minipage}{0.075\textwidth}
\includegraphics[width=1\textwidth, height=1.3cm]{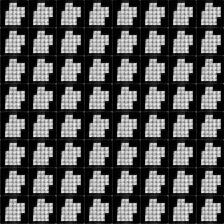}
\end{minipage}
& \begin{minipage}{0.075\textwidth}
\includegraphics[width=1\textwidth, height=1.3cm]{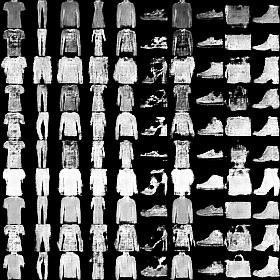}
\end{minipage}
& \begin{minipage}{0.075\textwidth}
\includegraphics[width=1\textwidth, height=1.3cm]{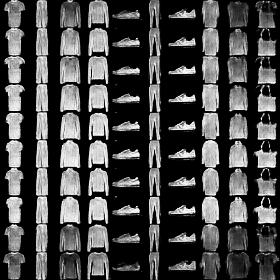}
\end{minipage}
& \begin{minipage}{0.075\textwidth}
\includegraphics[width=1\textwidth, height=1.3cm]{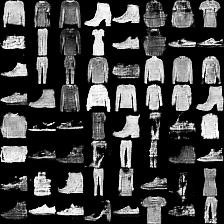}
\end{minipage}
& \begin{minipage}{0.075\textwidth}
\includegraphics[width=1\textwidth, height=1.3cm]{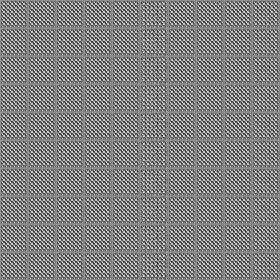}
\end{minipage} \\ \hline \hline
\begin{minipage}{0.075\textwidth}
\includegraphics[width=1\textwidth, height=1.3cm]{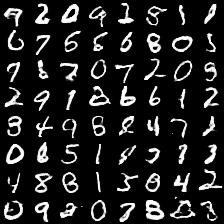}
\end{minipage}
& \begin{minipage}{0.075\textwidth}
\includegraphics[width=1\textwidth, height=1.3cm]{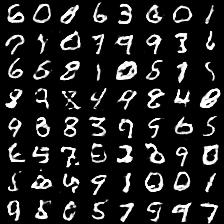}
\end{minipage}
& \begin{minipage}{0.075\textwidth}
\includegraphics[width=1\textwidth, height=1.3cm]{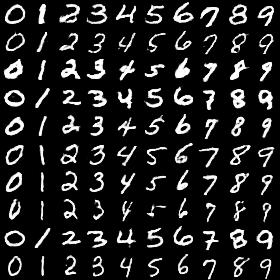}
\end{minipage}
& \begin{minipage}{0.075\textwidth}
\includegraphics[width=1\textwidth, height=1.3cm]{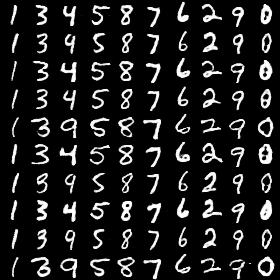}
\end{minipage}
& \begin{minipage}{0.075\textwidth}
\includegraphics[width=1\textwidth, height=1.3cm]{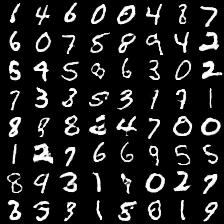}
\end{minipage}
& \begin{minipage}{0.075\textwidth}
\includegraphics[width=1\textwidth, height=1.3cm]{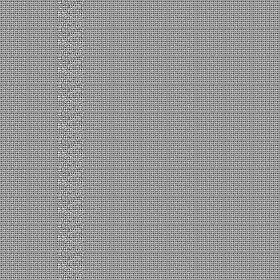}
\end{minipage}
& \begin{minipage}{0.075\textwidth}
\includegraphics[width=1\textwidth, height=1.3cm]{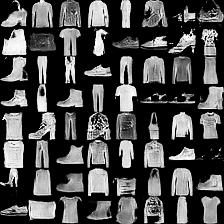}
\end{minipage}
& \begin{minipage}{0.075\textwidth}
\includegraphics[width=1\textwidth, height=1.3cm]{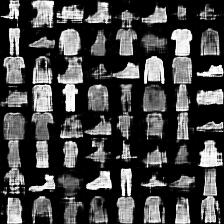}
\end{minipage}
& \begin{minipage}{0.075\textwidth}
\includegraphics[width=1\textwidth, height=1.3cm]{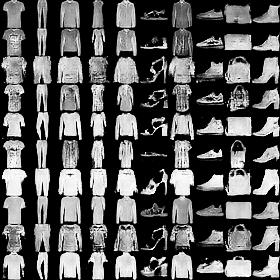}
\end{minipage}
& \begin{minipage}{0.075\textwidth}
\includegraphics[width=1\textwidth, height=1.3cm]{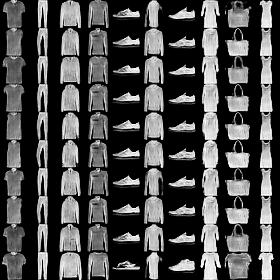}
\end{minipage}
& \begin{minipage}{0.075\textwidth}
\includegraphics[width=1\textwidth, height=1.3cm]{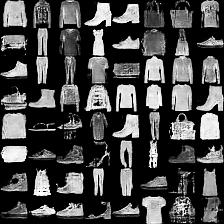}
\end{minipage}
& \begin{minipage}{0.075\textwidth}
\includegraphics[width=1\textwidth, height=1.3cm]{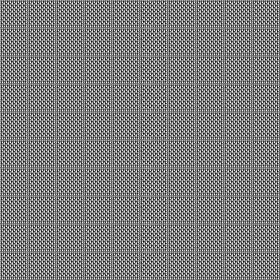}
\end{minipage} \\ \hline \hline
\begin{minipage}{0.075\textwidth}
\includegraphics[width=1\textwidth, height=1.3cm]{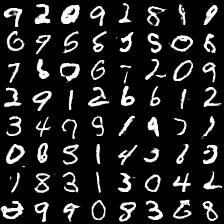}
\end{minipage}
& \begin{minipage}{0.075\textwidth}
\includegraphics[width=1\textwidth, height=1.3cm]{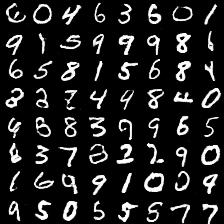}
\end{minipage}
& \begin{minipage}{0.075\textwidth}
\includegraphics[width=1\textwidth, height=1.3cm]{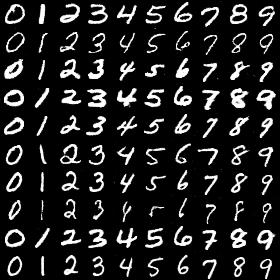}
\end{minipage}
& \begin{minipage}{0.075\textwidth}
\includegraphics[width=1\textwidth, height=1.3cm]{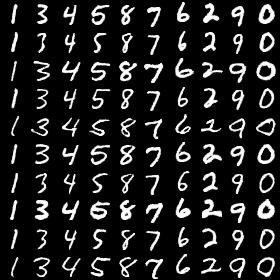}
\end{minipage}
& \begin{minipage}{0.075\textwidth}
\includegraphics[width=1\textwidth, height=1.3cm]{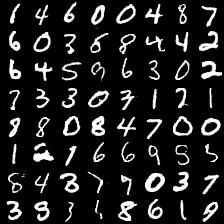}
\end{minipage}
& \begin{minipage}{0.075\textwidth}
\includegraphics[width=1\textwidth, height=1.3cm]{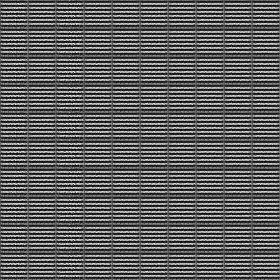}
\end{minipage}
& \begin{minipage}{0.075\textwidth}
\includegraphics[width=1\textwidth, height=1.3cm]{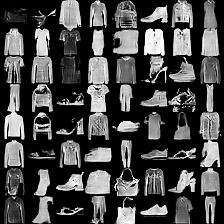}
\end{minipage}
& \begin{minipage}{0.075\textwidth}
\includegraphics[width=1\textwidth, height=1.3cm]{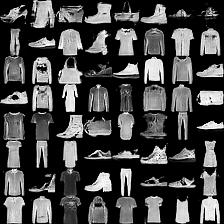}
\end{minipage}
& \begin{minipage}{0.075\textwidth}
\includegraphics[width=1\textwidth, height=1.3cm]{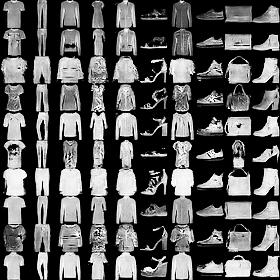}
\end{minipage}
& \begin{minipage}{0.075\textwidth}
\includegraphics[width=1\textwidth, height=1.3cm]{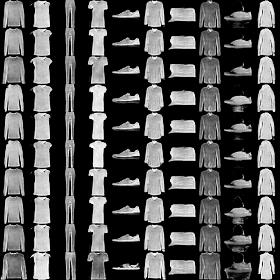}
\end{minipage}
& \begin{minipage}{0.075\textwidth}
\includegraphics[width=1\textwidth, height=1.3cm]{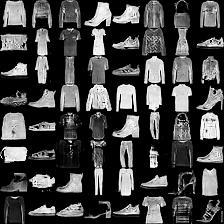}
\end{minipage}
& \begin{minipage}{0.075\textwidth}
\includegraphics[width=1\textwidth, height=1.3cm]{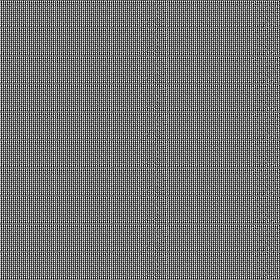}
\end{minipage} 
\end{tabular}
\vspace*{-1em}\caption{\small{Sample images generated by running all approaches for 1, 10, 25, 100 epochs.}}\label{Fig:sample_images_mnist}\vspace*{-1em}
\end{figure*}
\begin{figure*}[!t]\small
\hspace*{-5.5em}\begin{tabular}{ccc|ccc||ccc|ccc}
\multicolumn{6}{c}{\textsc{cifar10}} & \multicolumn{6}{c}{\textsc{imagenet}} \\ \hline
\multicolumn{3}{c}{DCGAN} & \multicolumn{3}{c}{MLP} & \multicolumn{3}{c}{DCGAN} & \multicolumn{3}{c}{MLP} \\ \hline
\footnotesize{RWGAN} & \footnotesize{WGAN} & \footnotesize{WGAN(g)} & \footnotesize{RWGAN} & \footnotesize{WGAN} & \footnotesize{WGAN(g)} & \footnotesize{RWGAN} & \footnotesize{WGAN} & \footnotesize{WGAN(g)} & \footnotesize{RWGAN} & \footnotesize{WGAN} & \footnotesize{WGAN(g)} \\
\begin{minipage}{0.075\textwidth}
\includegraphics[width=1\textwidth, height=1.3cm]{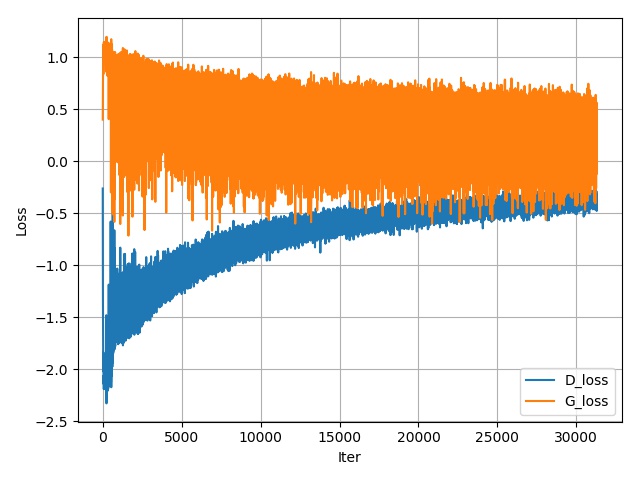}
\end{minipage}
& \begin{minipage}{0.075\textwidth}
\includegraphics[width=1\textwidth, height=1.3cm]{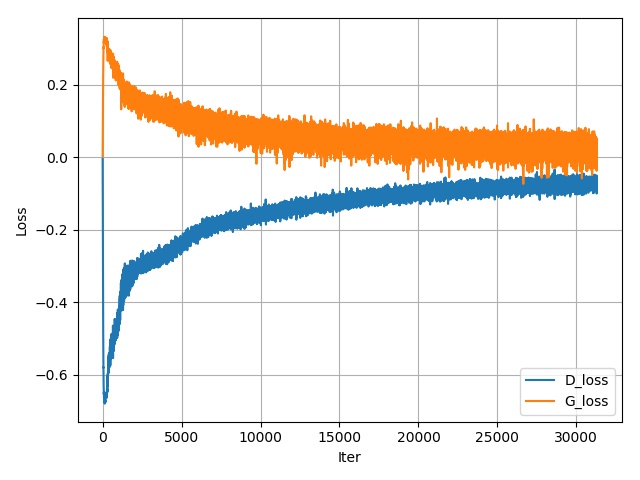}
\end{minipage}
& \begin{minipage}{0.075\textwidth}
\includegraphics[width=1\textwidth, height=1.3cm]{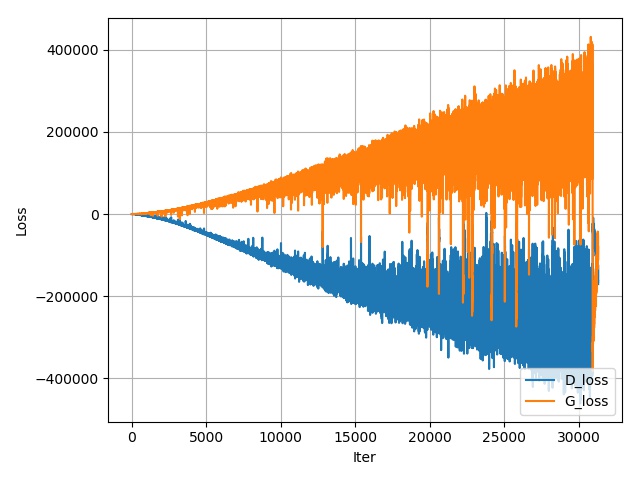}
\end{minipage}
& \begin{minipage}{0.075\textwidth}
\includegraphics[width=1\textwidth, height=1.3cm]{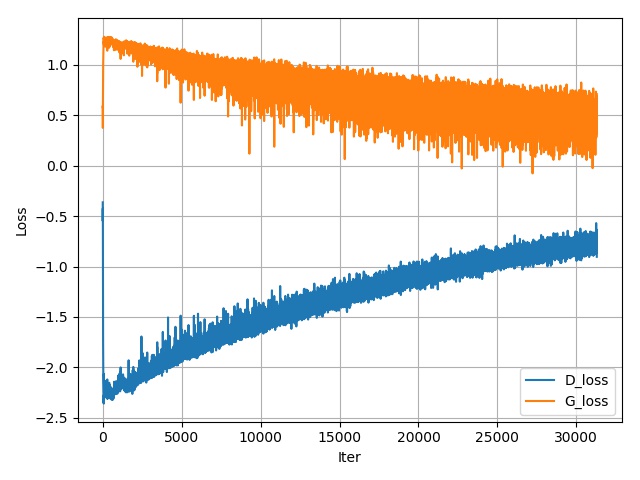}
\end{minipage}
& \begin{minipage}{0.075\textwidth}
\includegraphics[width=1\textwidth, height=1.3cm]{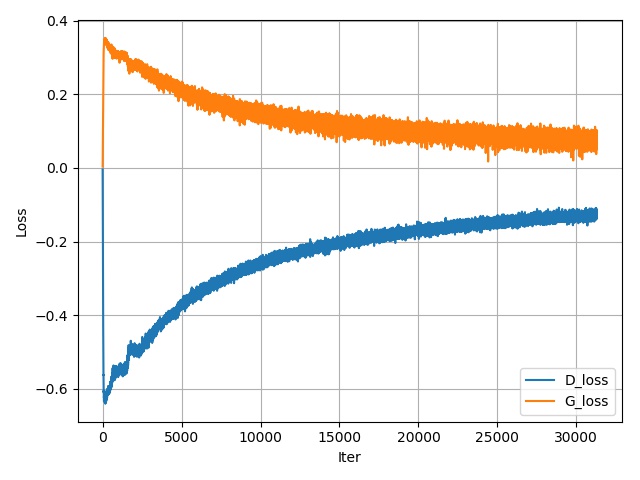}
\end{minipage}
& \begin{minipage}{0.075\textwidth}
\includegraphics[width=1\textwidth, height=1.3cm]{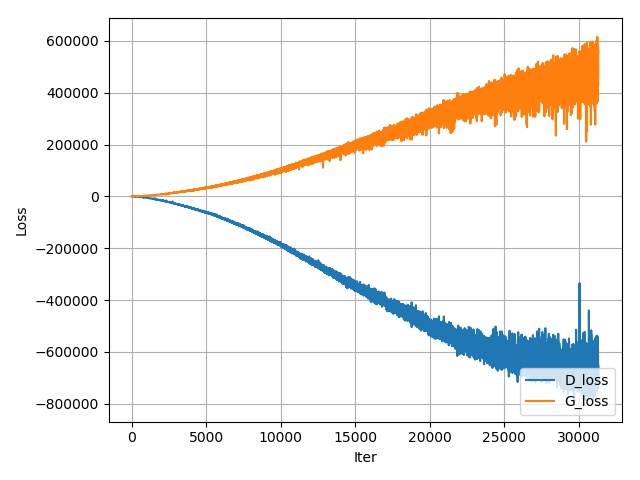}
\end{minipage}
& \begin{minipage}{0.075\textwidth}
\includegraphics[width=1\textwidth, height=1.3cm]{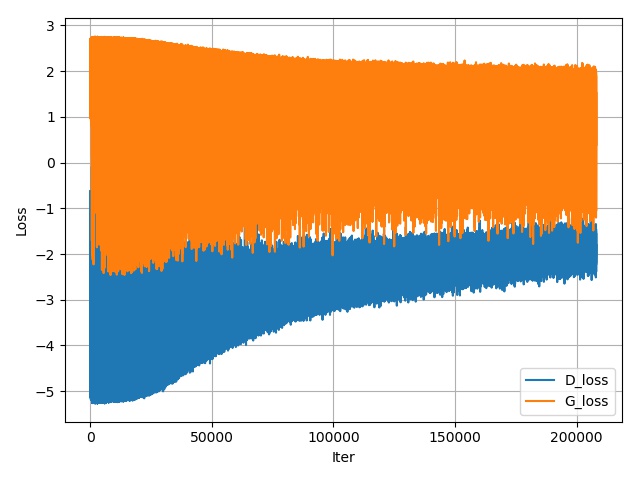}
\end{minipage}
& \begin{minipage}{0.075\textwidth}
\includegraphics[width=1\textwidth, height=1.3cm]{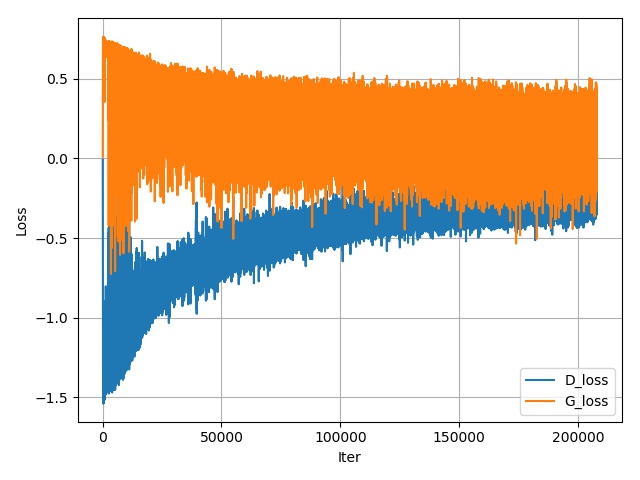}
\end{minipage}
& \begin{minipage}{0.075\textwidth}
\includegraphics[width=1\textwidth, height=1.3cm]{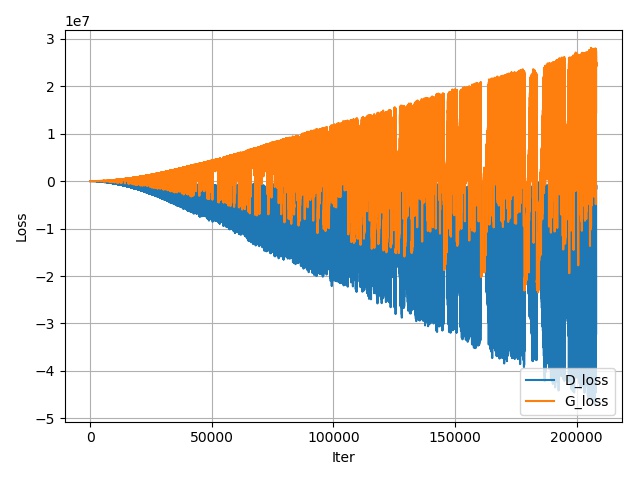}
\end{minipage}
& \begin{minipage}{0.075\textwidth}
\includegraphics[width=1\textwidth, height=1.3cm]{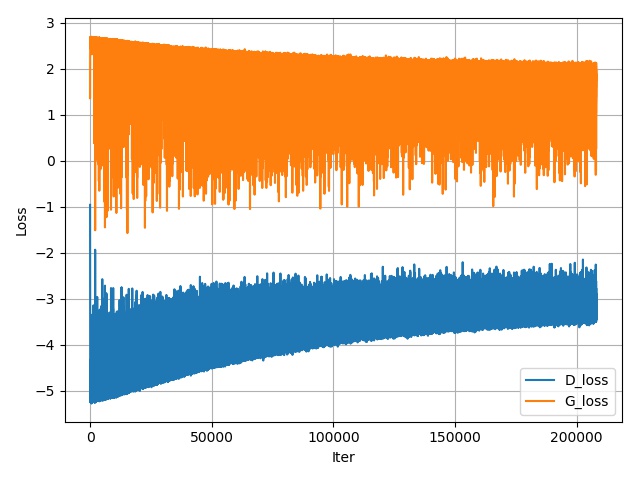}
\end{minipage}
& \begin{minipage}{0.075\textwidth}
\includegraphics[width=1\textwidth, height=1.3cm]{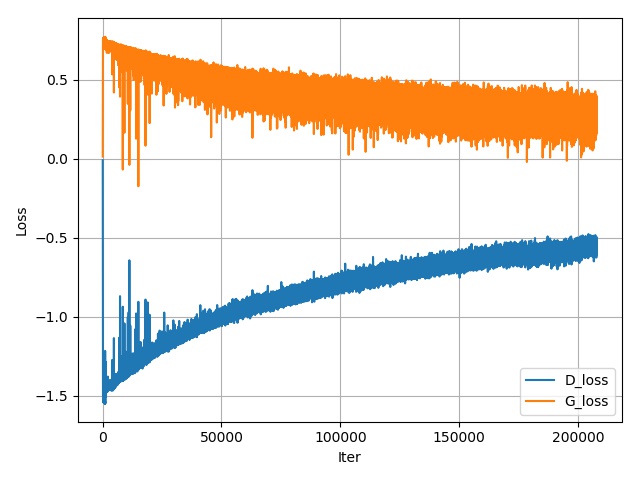}
\end{minipage}
& \begin{minipage}{0.075\textwidth}
\includegraphics[width=1\textwidth, height=1.3cm]{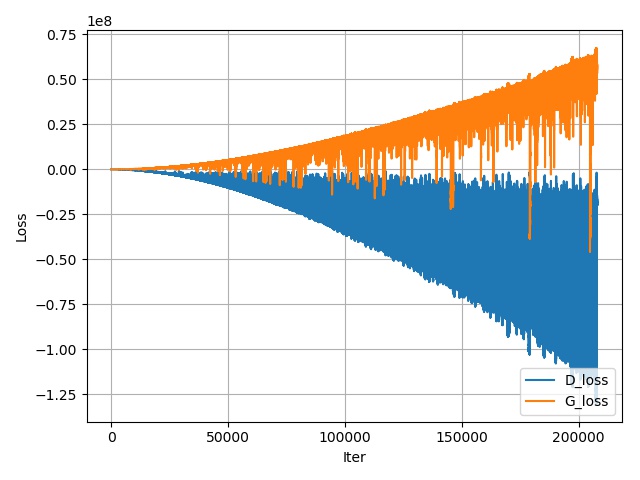}
\end{minipage}
\end{tabular}
\vspace*{-1em}\caption{\small{Training curves by running all approaches. Generator and discriminator losses are plotted in orange and blue lines.}}\label{Fig:train_curve_cifar10_imagenet}\vspace*{-1.2em}
\end{figure*}
\begin{figure*}[!t]\small
\hspace*{-5.5em}\begin{tabular}{ccc|ccc||ccc|ccc}
\multicolumn{6}{c}{\textsc{cifar10}} & \multicolumn{6}{c}{\textsc{imagenet}} \\ \hline
\multicolumn{3}{c}{DCGAN} & \multicolumn{3}{c}{MLP} & \multicolumn{3}{c}{DCGAN} & \multicolumn{3}{c}{MLP} \\ \hline
\footnotesize{RWGAN} & \footnotesize{WGAN} & \footnotesize{WGAN(g)} & \footnotesize{RWGAN} & \footnotesize{WGAN} & \footnotesize{WGAN(g)} & \footnotesize{RWGAN} & \footnotesize{WGAN} & \footnotesize{WGAN(g)} & \footnotesize{RWGAN} & \footnotesize{WGAN} & \footnotesize{WGAN(g)} \\
\begin{minipage}{0.075\textwidth}
\includegraphics[width=1\textwidth, height=1.3cm]{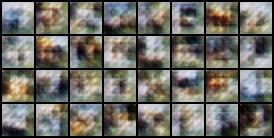}
\end{minipage}
& \begin{minipage}{0.075\textwidth}
\includegraphics[width=1\textwidth, height=1.3cm]{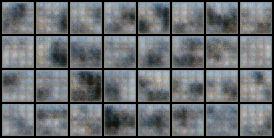}
\end{minipage}
& \begin{minipage}{0.075\textwidth}
\includegraphics[width=1\textwidth, height=1.3cm]{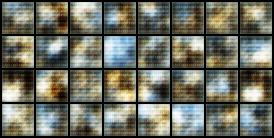}
\end{minipage}
& \begin{minipage}{0.075\textwidth}
\includegraphics[width=1\textwidth, height=1.3cm]{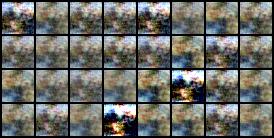}
\end{minipage}
& \begin{minipage}{0.075\textwidth}
\includegraphics[width=1\textwidth, height=1.3cm]{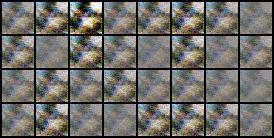}
\end{minipage}
& \begin{minipage}{0.075\textwidth}
\includegraphics[width=1\textwidth, height=1.3cm]{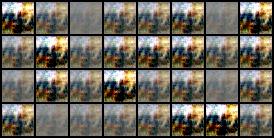}
\end{minipage}
& \begin{minipage}{0.075\textwidth}
\includegraphics[width=1\textwidth, height=1.3cm]{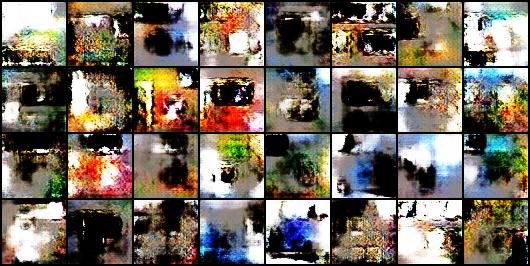}
\end{minipage}
& \begin{minipage}{0.075\textwidth}
\includegraphics[width=1\textwidth, height=1.3cm]{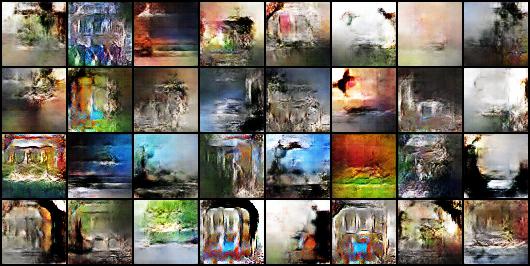}
\end{minipage}
& \begin{minipage}{0.075\textwidth}
\includegraphics[width=1\textwidth, height=1.3cm]{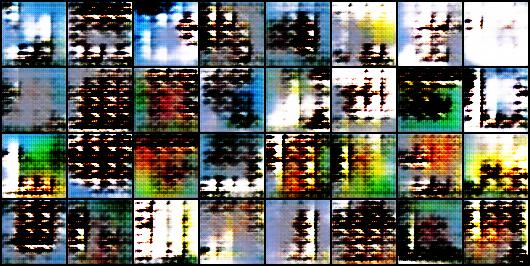}
\end{minipage}
& \begin{minipage}{0.075\textwidth}
\includegraphics[width=1\textwidth, height=1.3cm]{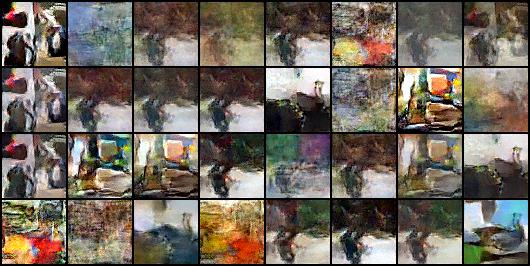}
\end{minipage}
& \begin{minipage}{0.075\textwidth}
\includegraphics[width=1\textwidth, height=1.3cm]{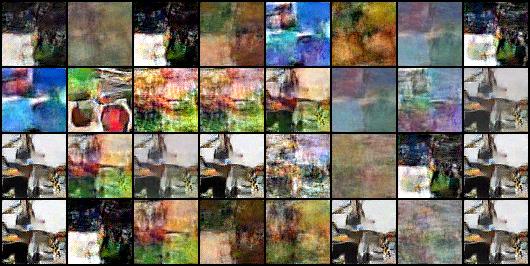}
\end{minipage}
& \begin{minipage}{0.075\textwidth}
\includegraphics[width=1\textwidth, height=1.3cm]{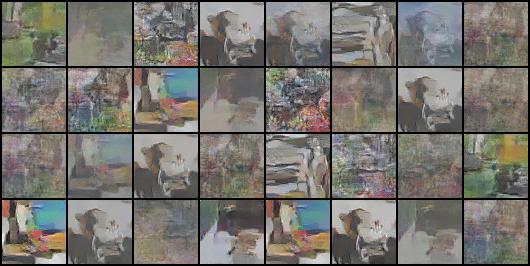} 
\end{minipage} \\ \hline \hline
\begin{minipage}{0.075\textwidth}
\includegraphics[width=1\textwidth, height=1.3cm]{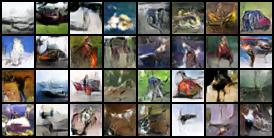}
\end{minipage}
& \begin{minipage}{0.075\textwidth}
\includegraphics[width=1\textwidth, height=1.3cm]{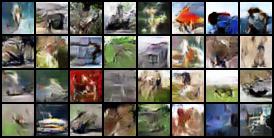}
\end{minipage}
& \begin{minipage}{0.075\textwidth}
\includegraphics[width=1\textwidth, height=1.3cm]{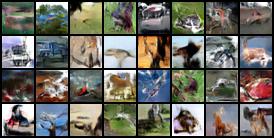}
\end{minipage}
& \begin{minipage}{0.075\textwidth}
\includegraphics[width=1\textwidth, height=1.3cm]{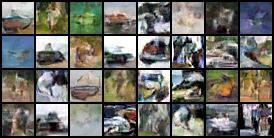}
\end{minipage}
& \begin{minipage}{0.075\textwidth}
\includegraphics[width=1\textwidth, height=1.3cm]{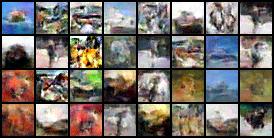}
\end{minipage}
& \begin{minipage}{0.075\textwidth}
\includegraphics[width=1\textwidth, height=1.3cm]{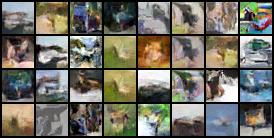}
\end{minipage}
& \begin{minipage}{0.075\textwidth}
\includegraphics[width=1\textwidth, height=1.3cm]{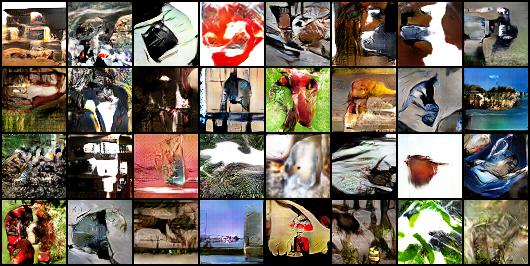}
\end{minipage}
& \begin{minipage}{0.075\textwidth}
\includegraphics[width=1\textwidth, height=1.3cm]{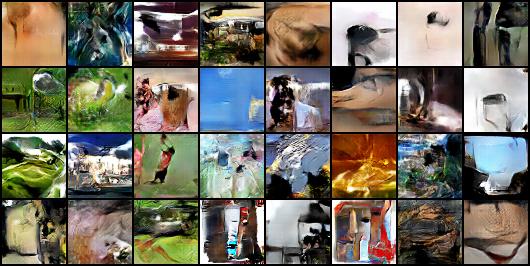}
\end{minipage}
& \begin{minipage}{0.075\textwidth}
\includegraphics[width=1\textwidth, height=1.3cm]{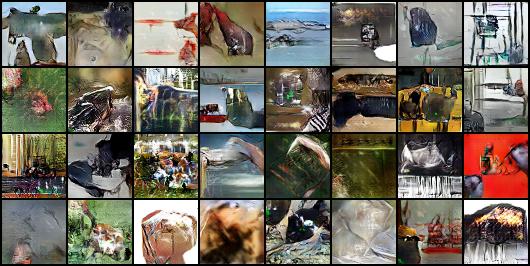}
\end{minipage}
& \begin{minipage}{0.075\textwidth}
\includegraphics[width=1\textwidth, height=1.3cm]{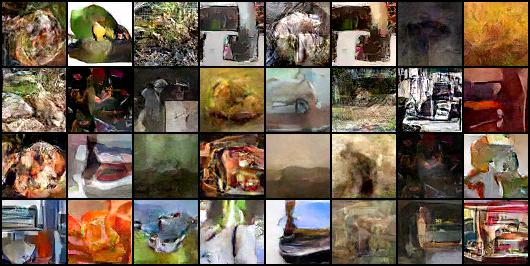}
\end{minipage}
& \begin{minipage}{0.075\textwidth}
\includegraphics[width=1\textwidth, height=1.3cm]{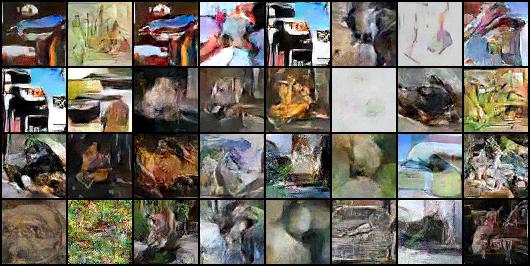}
\end{minipage}
& \begin{minipage}{0.075\textwidth}
\includegraphics[width=1\textwidth, height=1.3cm]{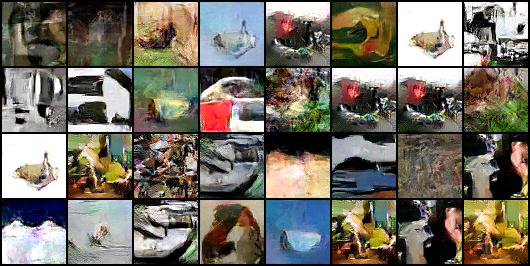}
\end{minipage}
\end{tabular}
\vspace*{-1em}\caption{\small{Sample images generated by running all approaches for 1, 100 (\textsc{cifar10}) or 25 (\textsc{imagenet}) epochs.}}\label{Fig:sample_images_cifar_imagenet}\vspace*{-1em}
\end{figure*}

\section{Experiments}\label{sec:experiments}
In this section, we report the numerical results which evaluate our approach on real images. All baseline approaches are discussed in the introduction and their implementations are available online\footnote{https://github.com/znxlwm/pytorch-generative-model-collections}. The datasets include \textsc{mnist}\footnote{http://yann.lecun.com/exdb/mnist/}, \textsc{fashion-mnist}\footnote{https://github.com/zalandoresearch/fashion-mnist/}, \textsc{cifar10}\footnote{https://www.cs.toronto.edu/$\sim$kriz/cifar.html} and \textsc{imagenet}\footnote{We use a small version of \textsc{imagenet} publicly available online: http://image-net.org/small/train{\_}64x64.tar}. The metrics include generator loss, discriminator loss and inception score~\cite{Saliman-2016-Improved}. While the former two ones characterize the stability of training, a high inception score stands for high quality of images generated by the model. 

For \textsc{mnist} and \textsc{fashion-mnist}, we set the maximum epoch number as $100$ and consider both generator and discriminator using a convolutional architecture (DCGAN)~\cite{Radford-2015-Unsupervised} in our approach. Figure~\ref{Fig:train_curve_mnist} shows that RWGAN and WGAN achieve the most stable training among all approaches. Compared to WGAN, RWGAN suffers from relatively high variance, which however improves the training process. As indicated by Figure~\ref{Fig:sample_images_mnist}, RWGAN is the fastest to generate interpretable images.

For \textsc{cifar10} and \textsc{imagenet}, besides the DCGAN architecture, we also consider the generator using ReLU-MLP~\cite{Conan-2002-Multi} with 4 layers and 512 units and set the maximum epoch number as 100 for \textsc{cifar10} and 25 for \textsc{imagenet}. Figure \ref{Fig:train_curve_cifar10_imagenet} shows that RWGAN achieves a good balance between robustness and efficiency while WGAN(g) are highly unstable. The effectiveness of RWGAN is also proven using inception score and sample images obtained by running all approaches; see Table~\ref{Tab:IS} and Figure~\ref{Fig:sample_images_cifar_imagenet}. 

\section{Conclusion}\label{sec:conclusion}
We propose new \textit{Relaxed Wasserstein} (RW) distances by generalizing Wasserstein-1 distance with Bregman cost functions. RW distances enjoy favorable statistical and computational properties, motivating RWGANs for learning generative models. Experiments on real images show that RWGANs with KL cost function achieves a good balance between robustness and efficiency. Future directions include a proper choice of $\phi$ in practice. 

\section{Acknowledgment}
We would like to thank Yiqing Lin for helpful comments on the duality theorem for RW distances. We are grateful to anonymous referees for constructive suggestions that improve the results of the paper.

% \vfill\pagebreak

\bibliographystyle{IEEEbib}
\bibliography{ref}

\end{document}